# Hierarchical Learning Algorithm for the Beta Basis Function Neural Network

**Habib DHAHRI** [(1)], **Mohamed Adel ALIMI** [(2)]

[(1)] *Meknassy secondary school BP W 9140, Tunisia*
*Tel. +216-97-473-672, Fax +216-76-645-790*

[(2)] *REGIM: REsearch Group of Intelligent Machines, University of Sfax, ENIS,*
*Department of Electrical BP W-3038, Sfax, Tunisia*
*Tel +216-4-274-088, Fax. +216-4-275-595*

Email: habib_dhahri@uk2.net , adel.alimi@ieee.org

*Abstract*– *The paper presents a two-level learning method for the design of the Beta Basis Function Neural Network BBFNN. A Genetic Algorithm is employed at the upper level to construct BBFNN, while the key learning parameters :the width, the centers and the Beta form are optimised using the gradient algorithm at the lower level. In order to demonstrate the effectiveness of this hierarchical learning algorithm HLABBFNN, we need to validate our algorithm for the approximation of non-linear function.*

*Keywords: Beta neural network, function approximation, genetic algorithm, gradient learning, Beta function.*

## 1. Introduction

Evolutionary Algorithms (EA) are stochastic search techniques, based on natural evolution, which have been extensively used in the last years in different fields, principally for optimisation [3] [6]. In these techniques, a population of possible solutions to a problem is evolved systematically with self-improvement as their primary objective. Common types of Evolutionary Algorithms are the Genetic Algorithms, the Evolutionary Strategies, and the Evolutionary Programming: While Evolutionary Computation is the generic term, which is used for the description of the whole field of algorithms and techniques inspired by the processes of natural evolution.

The EA [3][4][5][6][8] is used for designing RBF neural network topologies. A key advantage of using the EA as a neural network learning method is that it is capable of achieving optimal or near optimal network topology.

In this paper, we deal with the so-called Beta Basis Function Neural Network BBFNN [2] that represents an interesting alternative in which we can approximate any function .We propose a two-level learning hierarchy of construction BBFNN network based on the combined EA and the gradient descent algorithms. Because the determination of the network size is a difficult task, we use the evolutionary strategies for the construction of a parsimonious BBFNN. Since we define the topology of the network, we propose to refine the learning parameters by using the gradient method to improve the generalisation performance. The reduction of the number of a hidden unit, will be seen in simulation results

The remainder of this paper is organized as follows: Section 2 considers the structure of the Beta basis function neural network. Section 3 provides further details of the evolutionary algorithm. Section 4 consists of the details of the gradient method for the beta neural network. And the results obtained are illustrated and compared in section 5. Finally, section 6 summarizes the conclusion drawn from this study.

## 2. Beta basis function neural network

The Beta basis function neural network BBFNN, considered in this paper, consists of input layer, a hidden layer of Beta function and a linear output layer. When the structure of the BBFNN has a single output, it is defined in the form

$$f(x)=y=\sum_{i=1}^{k} w_i \beta_i(x, x_{ci}, d_i, p_i, q_i) \qquad (1)$$

In this form x is the network input vector, k is the number of nodes, $w_i$ is the synaptic weights, $x_{ci}$ is the center of Beta function $d_i$ is the width of the function, and $p_i$ and $q_i$ are the parameter forms of the Beta. Hence, the Beta form is defined as follows:

$$\beta(x)=\begin{cases}\left[\frac{(p+q)(x-x_c)}{dp}+1\right]^p \left[\frac{(p+q)(x_c-x)}{dq}+1\right]^q & \text{if } x\in\left]x_c-\frac{dp}{p+q}, x_c+\frac{dq}{p+q}\right[ \\ 0 & \text{else,}\end{cases} \qquad (2)$$

where $p > 0$, $q > 0$, $x_0$, $x$ are the real parameters, $x_0 < x_1$ and

$$x_c = \frac{px_1 + qx_0}{p + q} \quad (3)$$

The corresponding network structure is shown in Fig.1

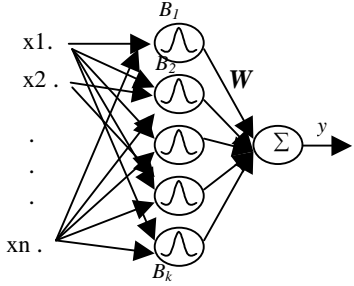

**Fig.1**. A structure of Beta neural network

Denote $T = \{x_i, d_i; i = 1. . . N\}$ a *training set*-a set of examples of network inputs $x_i \in \Re^n$ and desired outputs $d_i \in \Re$. For every training example, we can compute the actual network output $f(x_i)$ and error $e_i$ of the output units: $e_i = f(x_i) - d_i$. The goal of BBFNN network learning is to minimize the following error function, called *training error*:

$$E = \frac{1}{2} \sum_{j=1}^{K} e_j^2, \quad (4)$$

An important measure of a trained network performance is the so-called *generalization error*-an error (4) computed over a set of samples, which were not presented to the network during training. This set is referred to as a *testing set*.

## 3. The hierarchical algorithm for the training of Beta Neural Network

In this paper we present the hierarchical training algorithm based on a genetic algorithm (GA) and the gradient method. We adopt this hierarchical model (Fig.2): Genetic Algorithm at upper level to optimize the number of neuron in the hidden layer of BBFNN, because the GA is a powerful nonlinear optimization technique [9], and gradient method at the lower level to improve the generalization performance.

In this section, firstly we introduce the genetic algorithm (GA) then we pass to second technique such as the gradient.

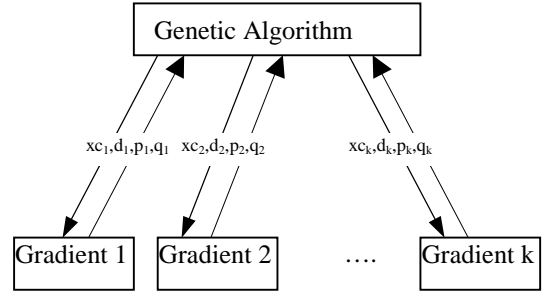

**Fig. 2.** Hierarchical Genetic Algorithm

### 3.1 Codifying the chromosomes

A generic chromosome consists of the four parameters $x_c$, d, p and q (fig. 3). The components $xc_i$, $d_i$, $p_i$, $q_i$, $1 \leq i \leq k$, are codified with the real numbers. Hence, we use a discrete representation to code the center $x_c$, the width d and the parameter forms p and q of every Beta function. Every chromosome that represents a network is a variable chain because the hidden layer contains a variable number of neurons. The number of parameters in the vector is variable and it depends on the number of neurons in the hidden layer of the network represented by this chromosome.

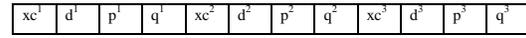

**Fig .3.** The encoding chromosome

The use of the same sequence in a chromosome several times does not improve the performance of the network. Therefore, every chromosome must contain distinct sequences.

The role of the Genetic Algorithm (GA) is the research of the optimal architecture of the BBFNN and these different parameters. Several operators have been used in order to change the lengths of the chromosomes that form the population to every generation. These operators are: the selection, reproduction, the crossover, the addition, the mutation and the elimination.

We choose the following objective function [2]

$$fitness1 = \sum_{i=1}^{i=K} (y_i - f(x_i))(y_i - f(x_i))^T, \quad (5)$$

$$fitness2 = (log(Nmax - nc + 1) + log(nc - Nmin + 1))\left[\frac{1}{1 + fitness1}\right]$$

, (6)

With Nmax, *Nmin* is respectively the maximal number and the minimal number of neurons in the hidden layer and *nc* is the number of neurons.

While looking at the architecture of the BBFNN network, we note that if we permute the position of two neurons in the hidden layer, the network output doesn't change. While being based on the previous idea, we can change the position of any neuron in the hidden layer. As every chromosome of the population is constructed by a set of sequences, each chromosome is constructed by the different parameters of every neuron; so we can permute the sequences of every chromosome without changing the output of the network itself.

### 3.2 The crossover operator

Let Pc be the probability of applying this operator. We get together the chromosomes at random in couples, and to every couple we affect a random real P chosen between 0 and 1. If P>Pc, then we are not going to apply the crossover on this couple. Therefore two chromosomes remain without changes. If the two chromosomes Chrom 1 and Chrom 2 (Fig. 4) have $l_1$ and $l_2$ respectively for length are chosen to apply the crossover on them, we choose $p_1$ an as arbitrary position in the first chromosome, then, we look for an arbitrary position $p_2$ in the second chromosome according to the position of $p_1$. In other words, if $p_1$ is between $x_c$ and d, then $p_2$ must be between $x_c$ and d.

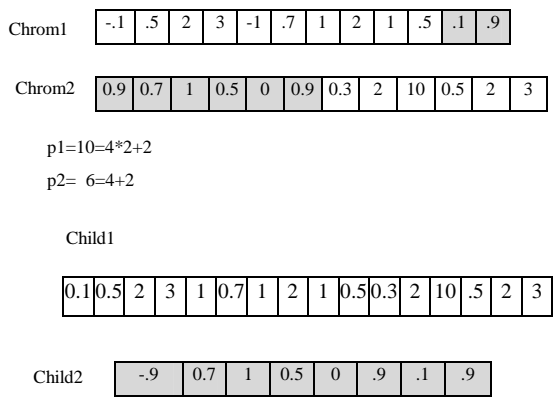

**Fig. 4.** The crossover of two chromosomes

Once, $p_1$ and $p_2$ are fixed, the first chromosome that is going to be constructed is called Chld2. This chromosome has $l_2 - p_2 + p_1$ for length and is composed as follows: The $p_1$ first term is the first term of Chrom1 and the $l_2 - p_2$ last terms are the $l_2 - p_2$ last terms of Chrom2. The second constructed chromosome is Chld1. This chromosome has $l_1 - p_1 + p_2$ for length and is composed as follows: The $p_2$ first terms are the $p_2$ first terms of Chrom2, whereas the $l_1 - p_1$ terms that remain are the $l_1 - p_1$ last terms of Chrom1. We know that the number of neurons in the hidden layer is therefore between Nmin and Nmax and that the lengths of the chromosomes must be between 4Nmin and 4Nmax. In this way, if Chld1 and Chld2 have some lengths between 4Nmin and 4Nmax, we pass to another couple of chromosomes of the population, which is going to be crossed. Otherwise, it is necessary to look for another $p_2$ position on Chrom2 as always taking into account the position $p_1$. If the stage described previously has been repeated K time and each time the length of Chld1 or Chld2 is not between 4Nmin and 4Nmax we pass to apply the second crossover operator.

### 3.3 The mutation operator

Generally, the initial population doesn't have all vital data for the optimal solution. For this reason, it is necessary to inject other information to the population in order to find a good solution. This injection is achieved by the mutation operator.

The intervention mutation operator makes itself as follows: Pm is the probability of mutation. For every term of the chromosomes that constitutes the population we associate arbitrariness real between 0 and 1. If the real value is lower than Pm, the mutation operator is applied and the value that is in this position is going to be replaced arbitrarily by another chosen value of the domain of definition of this value (Fig. 5).

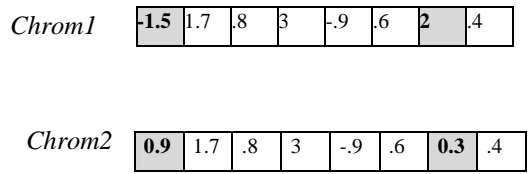

**Fig .5.** Mutation of two genes in the chromosome

### 3.4 The selection operator

The used selection operator is the original selection proposed by Holland [1] for the genetic algorithms. In this method of selection, the chromosomes are chosen with a proportional probability to their fitness. For example, an individual with a fitness of 2 will have twice as much luck as the one that has only one fitness. The metaphor of the roulette comes from the computer procedure to apply this proportional selection method. Every slot of the wheel is pondered in proportion to the value of fitness of every individual. Thus, the individuals having the biggest value of fitness will have more luck to be chosen.

### 3.5 The addition operator

We start with the selection for the chromosomes in which we apply this operator. This choice makes itself under the following way: We associate to every chromosome a real between 0 and 1, if this real is lower than the addition probability Pad, then the operator is applied to this chromosome. Once, the chromosome has been chosen, we apply on it the addition operator, we add to it a sequence that represents the four parameters of a Beta function. Therefore, it comes back to say that we added a new neuron in the layer hidden of the network represented by the chromosome (Fig .6).

| -.9 | .7 | 1 | .5 | 0 | .9 | .3 | 2 |

A chromosome before the applying of the addition operator

| -.9 | .7 | 1 | .5 | 0 | .9 | .3 | 2 | 1 | .5 | .1 | .9 |

The same chromosome after the applying of the addition operator

**Fig. 6.** Addition operator

### 3.6 The operator elimination

For the elimination operator that is applied with a Pel probability, we start with the selection of the chromosomes in which we apply this operator (Fig 7). It comes back therefore to look for the networks that we are going to decrease their numbers of neurons in their hidden layers by one. If a chromosome has been chosen, we eliminate any sequence of the chromosome in an arbitrary way.

| -.1 | .5 | 2 | 3 | -1 | .7 | 1 | 2 | 1 | .5 | .1 | .9 |

The chromosome before the applying of the elimination operator

| -.1 | .5 | 2 | 3 | 1 | .5 | .1 | .9 |

The same chromosome after the applying of the elimination operator .

**Fig.7.** The elimination operator

### 3.7. A learning algorithm based on Gradient descent for the Beta neural network.

After the determination of the number of neuron, the BBFNN neural network can be trained by minimizing the training error $E$. The updating equation for the learning parameters using gradient descent[7], the synaptic weights $w_i$, the centers $x_c$, the parameters form p and q for all the hidden nodes defined as follows

$$\Delta x_c = -\eta \frac{\partial E}{\partial x_c}, \qquad (7)$$

$$\Delta d = -\eta \frac{\partial E}{\partial d}, \qquad (8)$$

$$\Delta p = -\eta \frac{\partial E}{\partial p}, \qquad (9)$$

$$\Delta q = -\eta \frac{\partial E}{\partial q}, \qquad (10)$$

$$\Delta w_i = -\eta \frac{\partial E}{\partial w_i}, \qquad (11)$$

Where $\eta$ is a learning constant.

$$\Delta w_i = -\eta \frac{\partial E}{\partial y} \frac{\partial y}{\partial w_i}, \qquad (12)$$

$$\Delta w_i = \eta (t_i - y_i) \beta_i, \qquad (13)$$

$$\Delta x_{ci} = \eta \beta_i (p_i + q_i)(q_i A_i + p_i B_i)(t_i - y_i) w_i, \qquad (14)$$

$$\Delta d_i = \eta \beta_i \frac{(x - x_{ci})}{d_i}(p_i + q_i)(q_i A_i - p_i B_i)(t_i - y_i) w_i, \qquad (15)$$

$$\Delta p_i = \eta \beta_i \left( ln\left(\frac{1}{B_i p_i d_i}\right) - q_i(x - x_{ci})(A_i + B_i) \right) w_i (t_i - y_i), \qquad (16)$$

$$\Delta q_i = \eta \beta_i \left( ln\left(\frac{1}{A_i q_i d_i}\right) + p_i(x - x_{ci})(A_i + B_i) \right) w_i (t_i - y_i), \qquad (17)$$

Where

$$A_i = A_i(x) = \frac{1}{q_i d_i - (p_i + q_i)(x - x_{ci})}, \qquad (18)$$

$$B_i = B_i(x) = \frac{1}{p_i d_i + (p_i + q_i)(x - x_{ci})}. \qquad (19)$$

## 4. The algorithm of the training of the BBFNN

Step (1): Initialization.
Step (2): Choose randomly the initial population.
Step (3): Decode each chromosome.
Step (4): Compute the connection weight.
Step (5): Find the fitness *fitness2* of each chromosome (best chromosome). If the number of generations is completed then go to step (i) else go to step (6)
    Step (i): Update network parameter by equations (13), (14), (15), (16), (17).
    Step (ii): If the number of iteration is equal to maxiter (maximum iteration of gradient algorithm) then go to step (6), else if the learning error is less than err then go out.

Step (6): Apply the GA operator (, selection, crossover, mutation, addition and elimination operator) then go to step (3).

## 5. Experimental results

To prove the efficiency of our Hierarchical Learning Algorithm for the Beta Basis Function Neural Network (HLABBFNN), we use an example to test it in approximation function. The simulations were conducted in the Matlab environment. The gotten results are compared with the results obtained by another genetic algorithm of training of the networks of neurons to function of basis beta achieved by Aouiti [2]. So, the proposed function is the following:

$$g_2(x) = 10 \arctg\left[\frac{(x-0.2)(x-0.7)(x+0.8)}{(x+1.4)}\right], \quad (20)$$

In this example, the minimal number and the maximal number of functions Beta in our HLABBFNN algorithm are respectively 5 and 20, the number of individuals by population is equal to 50, the number of maximal generation is equal to 100, the stop criteria is equal to 0.01. The parameters of each beta function $x_c$, $d$, $p$, $q$ are respectively in [-1 1], [-1 1], [0,4], and [0 4].
We used 201 points for the training data and 126 points for the generalisation.
Figure 9 represents the output and the function desired proposed by HLABBFNN.

In this example, the result proposed by HLABBFNN For the design of the Beta basis function neural Network is better (Tab.1) than the technique in [2].

|  | RGADBBFNN | HLABBFNN |
|---|---|---|
| Training error | 0.073 | 0.0068 |
| Number neuron | 14 | 8 |
| Generation error | -- | 0.0069 |

**Tab.1** Comparison between the HGABBFNN and RGADBBFNN.

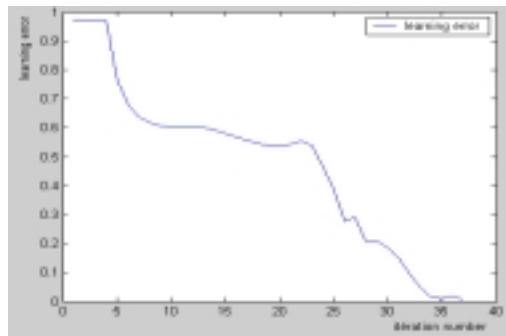

**Fig 8**. The training error

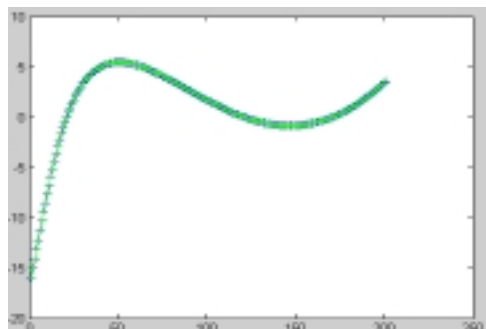

**Fig 9.** The output of HLGABBFNN (--) and the desired output (**)

## Conclusion

In this paper, two-level learning hierarchy has been developed by combining the GA with the gradient learning to improve Beta Basis neural Network BBFNN for the approximation problem. The GA is used at the upper level to find the optimal architecture of the Beta system among possible solutions. The learning parameters are optimized at the lower level by the user of the gradient approach. The proposed method is computationally more efficient compared with the one developed in [2].

## References

[1] C., Aouiti, Alimi, A. M., Karray, F. and Maalej, A. (2000). Beta Basis Funtion Neural Network: Approximation Properties, IEEE International conference on fuzzy Systems.
[2] C. Aouiti, A. M. Alimi, and A. Maalej (2001). A Genetic Design of Beta Basis Function Neural Network for approximating Multivariable functions ", V. Kurkova et al. Artificial Neural Nets and Genetic Algorithm, Springer-Verlag,Wien,pp.383-386.

[3] D. E. Goldberg (1989), Genetic Algorithms in Search, Optimisation, and Machine learning. Reading, MA: Addison-Wesley.

[4] D. Dumitrescu and R. Gorunescu (2000) A new dynamic evolutionary clustering technique. Application in designing RBF neural network topologies in clustering algorithm. Mathematics Subject Classification. 68T05, 68T20, 91C20, 92B20.

[5] D., Whitey and Bogard, C., (1990), The evolution of connectiviy: Pruning neural Network using Genetic Algorithms, in int. Joint Conf. Neural Network, Vol.1, pp. 134-137.

[6] J.H.Holland (1975), Adaptation in natural and artificial systems. Ann Arbor, MI: Univ .Michigan .

[7] M., Njah, M., Alimi, M. A., M., Chtourou, and T., R., (2002), Algorithm of Maximal Descent AMD for training Radial Basis Function Neural Network, Proceedings IEEE Confenrence on Systems Man and Cybernetics, SMC-02, Hammamet, Tunisia, tp1p1.

[8] Whitehead, B.A., (1995), Genetic evolution of radial basis function coverage using orthogonal niches, IEEE Transact.on neural Netw.,vol. 7, n °5, pp. 1525-1558.

[9] S.A. Billings and G. L. Zheng, (1196), Radial basis function network configuration using genetic algorithms, Neural Network, vol. 8, no. 6 pp.877-890.